\documentclass[letterpaper]{IEEEtran}
\usepackage{latexsym,,amssymb,amsmath,graphicx,epsf,cite,bbm,float}
\usepackage{ifpdf}
\usepackage{epstopdf}

\usepackage{algorithm,algorithmic}
\usepackage{amsmath,amssymb,bm}
\usepackage{amsfonts,dsfont,color,bbm,subcaption}

\usepackage[linguistics]{forest}

\newcommand{\be}{\begin{equation}}
\newcommand{\ee}{\end{equation}}
\newcommand{\bea}{\begin{eqnarray}}
\newcommand{\eea}{\end{eqnarray}}

\newcommand{\MB}{\left[\begin{array}}
\newcommand{\ME}{\end{array}\right]}

\newcommand{\ei}{\end{itemize}}
\newcommand{\bi}{\begin{itemize}}

\newcommand{\X}{\mathcal{X}}
\newcommand{\Y}{\mathcal{Y}}
\newcommand{\Z}{\mathcal{Z}}

\newcommand{\A}{\mathcal{A}}

\usepackage{amsmath,amsthm}
\usepackage{hyperref}

\DeclareMathOperator*{\argmin}{arg\,min}

\newtheorem{theorem}{Theorem}

\newtheorem{lemma}[]{Lemma}

\newtheorem{proposition}[]{Proposition}

\newtheorem{remark}[]{Remark}

\newtheorem{definition}[]{Definition}

\begin{document}

\title{A Log-Linear Time Sequential Optimal Calibration Algorithm for Quantized Isotonic L2 Regression} 
\author{
	\IEEEauthorblockN{Kaan Gokcesu}, \IEEEauthorblockN{Hakan Gokcesu}
}
\maketitle

\flushbottom
\begin{abstract}
	We study the sequential calibration of estimations in a quantized isotonic L2 regression setting. We start by showing that the optimal calibrated quantized estimations can be acquired from the traditional isotonic L2 regression solution. We modify the traditional PAVA algorithm to create calibrators for both batch and sequential optimization of the quantized isotonic regression problem. Our algorithm can update the optimal quantized monotone mapping for the samples observed so far in linear space and logarithmic time per new unordered sample.

\end{abstract}

\section{Introduction}

	The minimization of certain error or loss functions are heavily studied in  the problems of machine learning, prediction and estimation \cite{poor_book, cesa_book,russel2010}; especially in the applications pertaining to control theory \cite{tnnls3}, optimization \cite{zinkevich, hazan}, scheduling \cite{kose2020novel}, forecasting \cite{singer, gokcesu2016prediction}, decision theory \cite{tnnls4}, density estimation and anomaly detection \cite{gokcesu2018density, willems, gokcesu2018anomaly,coding2,gokcesu2019outlier, gokcesu2016nested}, bandits \cite{neyshabouri2018asymptotically,cesa-bianchi,gokcesu2018bandit}, signal processing \cite{ozkan,gokcesu2018semg} and game theory \cite{tnnls1, chang}.
	In most of the applications, a learner or decision maker produces its estimates in the form of some score. Even though, for some applications, the ordinality implied by these estimates are sufficient; calibration is needed for increased accuracy, which is generally achieved with post-processing calibrators \cite{lichtenstein1977calibration,keren1991calibration}. 
	Although calibrated outputs generally perform better, its unlikely to achieve perfect matching \cite{naeini2015obtaining}. Nonetheless, well-calibrated predictions are important in many heavily studied fields such as business and medicine. These well-calibrated estimates are used not only to increase decision making performance but also for more accurate evaluation of learning models \cite{zhang2004naive, hashemi2010application} and even their more effective ensemble \cite{bella2013effect}. While lacking in comparison to the effort in creating models with strong discriminative performance, calibration is a rich field of study that includes many established post-processing methods that utilize some regularization (against over-fit) \cite{gokcesu2021optimally}. 
	
	The most straightforward calibration technique regularizes the output estimations by mapping the inputs using an optimized parametric function (e.g., a step function with optimized threshold for binary classification \cite{gokcesu2021optimally}). The most widely used technique involves a sigmoidal fit \cite{platt1999probabilistic}, where the input estimations are passed through a logistic function with appropriately tuned parameters \cite{gill2019practical}.
	It has been empirically shown that this parametric technique performs satisfactorily despite its efficiency in contrast to training another learning model \cite{platt1999probabilistic}. It has been successfully utilized in the calibration of many different learning algorithms \cite{niculescu2005predicting,bennett2000assessing} in addition to the original study on SVM \cite{platt1999probabilistic}. However, in another empirical study, it has also been shown that the sigmoidal fitting technique may under-perform for different learning models \cite{zadrozny2002transforming}.
	The unsatisfactory performance arises because of the strong restriction imposed by the parametric shape \cite{jiang2012calibrating}.
	
	The restrictiveness of the parametric calibration is relaxed by the binning approach \cite{zadrozny2001learning, zadrozny2001obtaining}, which starts by partitioning the ordered input estimations in bins. The output estimations are created from the respective bins individually, which imposes the regularization of mapping similar estimations to similar outputs. 
	Because of its less restrictive nature in contrast to the sigmoidal fitting, its parameters (e.g., the bin sizes and locations) are harder to optimize \cite{zadrozny2002transforming}. Incorrect determination of the bin parameters may lead to ineffective learning. To efficiently address this issue, confidence intervals are utilized to create the bins \cite{jiang2012calibrating}. To address the inherent limitations of the binning approach, ensemble methods that use Bayesian mixture \cite{heckerman1995learning} are studied \cite{naeini2015obtaining,naeini2015binary}. Although the relaxed restriction of binning provides more freedom in the design, it may lead to over-fitting.
	
	In the field of calibration, the inherent drawbacks of the sigmoid fitting and quantile binning approaches are addressed by the technique of isotonic regression \cite{robertson1988order,zadrozny2002transforming,menon2012predicting}, whose regularity originates from the monotonicity on the mapping, i.e., the calibrated estimations share the same ordinality with the input estimations. This regularity results in the mapping of several adjacent samples to the same calibrated value (hence the name isotonic). The number of samples mapped to a single value increases as the input estimates are incorrectly ranked (bad discriminative performance).
	From the perspective of restrictiveness, it is between sigmoid fitting and binning; and shares certain similarities with them \cite{zadrozny2002transforming,gokcesu2021optimally}. Many algorithms were designed to create isotonic regressors \cite{ayer1955empirical,brunk1972statistical}. It has been utilized to help quantile binning strategies \cite{naeini2016binary,tibshirani2011nearly}; and to 
	aggregate distinct input estimates \cite{zhong2013accurate}. 	

Although the isotonic regression addresses many issues with the sigmoid fitting and quantile binning, it is traditionally hard to update \cite{gokcesu2021optimally}. Even though a sequential algorithm for isotonic mapping can have up to quadratic time complexity in general \cite{gokcesu2021efficient}, we show that when focused on the problem of quantized regression, a log-linear complexity algorithm exists. To this end, we propose an efficient sequential algorithm with log-linear time and linear space complexity for the problem of quantized isotonic L2 regression.

In \autoref{sec:prob}, we formally define the quantized isotonic L2 regression. In \autoref{sec:optimal}, we prove that an optimal quantized monotone mapping can be extracted from an optimal unconstrained monotone mapping. In \autoref{sec:prefix}, we design an algorithm that can sequentially update the optimal quantized monotone mapping in logarithmic time with each new sample when the scores arrive in order. In \autoref{sec:recursive}, we propose a sequential algorithm that can update the optimal quantized monotone mapping in logarithmic time with each new sample even when the samples scores are observed out of order.

\section{Problem Definition}\label{sec:prob}
	For $N$ number of samples indexed by $n\in\{1,2,\ldots,N\}$, we have the target variables $y_n$ where
	\begin{align}
		y_n\in\Re,\label{eq:y}
	\end{align}
	where $\Re$ is the set of reals. For every $y_n$, we have unordered input estimations $x_n$ (i.e., the score of the $n^{th}$ sample) where
	\begin{align}
		x_n\in \Re.\label{eq:x}
	\end{align} 
	Moreover, we have the positive sample weights $\alpha_n$ where
	\begin{align}
		\alpha_n>0.\label{eq:a}
	\end{align}
	Thus, the initial uncalibrated loss is given by
	\begin{align}
		\sum_{n=1}^{N}\alpha_n(y_n-x_n)^2.
	\end{align}
	In the isotonic regression problem, the traditional goal is to create calibrated estimations $z_n\in \Re$ to decrease the weighted square error, where
	\begin{align}
		z_n=C(x_n),
	\end{align}
	for some monotone transform $C(\cdot)$ such that
	\begin{align}
		C(x_{n})\leq C(x_{n'})  &&\text{if } x_n\leq x_{n'} &&\forall n,n'.
	\end{align}
	\begin{definition}
		The objective of the traditional isotonic L2 regression problem is given by
		\begin{align}
			\argmin_{C(\cdot)\in\Omega}\sum_{n=1}^{N}\alpha_n(y_n-C(x_n))^2,\label{eq:objori}
		\end{align}
		where $\Omega$ is the class of all monotonically nondecreasing functions that map from $\Re$ to $\Re$.
	\end{definition}
	In this paper, we study the quantized version of this isotonic regression problem, where
	\begin{align}
		z_n=C_Q(x_n)\in \mathcal{Q},
	\end{align}
	for some quantized subset $\mathcal{Q}\subset \Re$.
	\begin{definition}\label{def:isoquant}
		The revised objective of the quantized isotonic L2 regression is given by
		\begin{align}
			\argmin_{C_Q(\cdot)\in\Omega_Q}\sum_{n=1}^{N}\alpha_n(y_n-C_Q(x_n))^2,\label{eq:objrev}
		\end{align}
		where $\Omega_Q$ is the class of all monotonically nondecreasing functions that map from $\Re$ to a quantization subset $\mathcal{Q}$.
	\end{definition}
	
	Let an optimizer for the objective in \autoref{def:isoquant} be $C_N(\cdot)$ for $N$ number of samples. Our goal is to sequentially create these optimizers in an efficient manner. After we acquire $C_N(\cdot)$, we want to update the mapping and create $C_{N+1}(\cdot)$ using the new sample $x_{N+1}$. We will show that we can update the mapping in a time complexity that is logarithmic in the number of samples observed so far.

	
	\section{From an Optimal Monotone Mapping to an Optimal Quantized Monotone Mapping }\label{sec:optimal}
	Let us consider that the sample scores $x_n$ are ordered, i.e., $x_n\leq x_{n+1}$ for all $n$ (if not we can just order them).
	Let us have a minimizer $C_N(\cdot)$ for \autoref{def:isoquant} given by
	\begin{align}
		C_N(x_n)=z_n, &&n\in\{1,\ldots,N\},\label{eq:prob}
	\end{align} 
	where $z_n$ be the corresponding mapped value.
	
	\subsection{Optimal Quantized Monotone Mapping}
	\begin{lemma}\label{thm:sample}
		There exists a minimizer $C_N(\cdot)$ for \autoref{def:isoquant} such that
		\begin{align*}
			z_{n}=z_{n+1}\text{ if }y_n\geq y_{n+1}
		\end{align*}
	\begin{proof}
		Let us have an initial quantized monotone mapping $\{z'_n\}_{n=1}^N$, where $z'_n\in\mathcal{Q}$ and $z'_n\leq z'_{n+1}$ for all $n$. Let the loss of interest be $L_n(z'_n,z'_{n+1})=\alpha_n(z'_n-y_n)^2+\alpha_{n+1}(z'_{n+1}-y_{n+1})^2$. Given $y_n\geq y_{n+1}$, we have the following six cases:
		\begin{enumerate}
			\item $z'_n\leq z'_{n+1}\leq y_{n+1}\leq y_n$:
			\subitem $L_n(z'_{n+1},z'_{n+1})\leq L_n(z'_{n},z'_{n+1})$
			\item $z'_n\leq y_{n+1}\leq z'_{n+1}\leq y_n$:
			\subitem $L_n(z'_{n+1},z'_{n+1})\leq L_n(z'_{n},z'_{n+1})$.
			\item $ z'_n\leq y_{n+1}\leq y_n\leq z'_{n+1}$:
			\subitem $L_n(\theta_n,\theta_n)\leq L_n(z'_{n},z'_{n+1})$, $\exists\theta_n\in[z'_{n},z'_{n+1}]\cap \mathcal{Q}$.
			\item $y_{n+1}\leq z'_n\leq z'_{n+1}\leq  y_n$:
			\subitem $L_n(\theta_n,\theta_n)\leq L_n(z'_{n},z'_{n+1})$, $\forall\theta_n\in[z'_{n},z'_{n+1}]\cap \mathcal{Q}$.
			\item $y_{n+1}\leq z'_n\leq y_n\leq z'_{n+1}$:
			\subitem $L_n(z'_{n},z'_{n})\leq L_n(z'_{n},z'_{n+1})$.
			\item $y_{n+1}\leq y_n\leq z'_n\leq z'_{n+1}$:
			\subitem $L_n(z'_{n},z'_{n})\leq L_n(z'_{n},z'_{n+1})$.
		\end{enumerate}
		Therefore, there exists a minimizer $C_N(\cdot)$ such that $z_{n}=z_{n+1}$ if $y_n\geq y_{n+1}$, which concludes the proof.
	\end{proof}
	\end{lemma}

	Hence, there are groups of samples with the same quantized mapping $z_n\in\mathcal{Q}$. Following \cite{gokcesu2021optimally,gokcesu2021efficient}, let there be $I$ such groups, where the group $i\in\{1,\ldots,I\}$ consists of the samples $n\in\mathcal{N}_i=\{n_i+1,\ldots,n_{i+1}\}$ (with $n_1=0$, $n_{I+1}=N$), i.e., for each group $i$, we have $z_n=z_{n'}, \forall n,n'\in\mathcal{N}_i$. With an abuse of notation, let us denote the mapping of the $i^{th}$ group with $z_i$. We have the following result.
	
	\begin{lemma}\label{thm:group}
	If $C_N(\cdot)$ is a minimizer for \autoref{def:isoquant}, then
	\begin{align}
		z_{i}=z_{i+1}\text{ if }\tilde{y}_i\geq \tilde{y}_{i+1},
	\end{align}
	where $\tilde{y}_i$ is the weighted average of the target variables $y_n$ of group $i$, i.e., 
	\begin{align*}
		\tilde{y}_i\triangleq \frac{\sum_{n=n_i+1}^{n_{i+1}}\alpha_ny_{n}}{\sum_{n=n_i+1}^{n_{i+1}}\alpha_n}.
	\end{align*}
	Consequently, we have
	\begin{align}
		z_{i}<z_{i+1} \text{ only if } \tilde{y}_i<\tilde{y}_{i+1},
	\end{align}
	i.e., the $i^{th}$ groups mapping $z_i$ is distinct (consequently strictly less) than $z_{i+1}$ only if $\tilde{y}_i$ is strictly less than $\tilde{y}_{i+1}$.
	\begin{proof}
		The proof follows similar arguments with the proof of \autoref{thm:sample}.
	\end{proof}
	\end{lemma}
	
	\subsection{Achieving Optimal Quantized Monotone Mapping}
	\begin{proposition}\label{thm:quant}
		We have the following quantization result
		\begin{align*}
			\argmin_{z\in\mathcal Q}\sum_{n=n_i+1}^{n_{i+1}}\alpha_n(z-y_n)^2=\argmin_{z\in\mathcal{Q}}\left|z-\frac{\sum_{n=n_i+1}^{n_{i+1}}\alpha_ny_{n}}{\sum_{n=n_i+1}^{n_{i+1}}\alpha_n}\right|
		\end{align*}
		\begin{proof}
		Let $\tilde{y}_i$ be the weighted average. We have
		\begin{align}
			\sum_{n=n_i+1}^{n_{i+1}}\alpha_n(z-y_n)^2=&\sum_{n=n_i+1}^{n_{i+1}}\alpha_n(z-\tilde{y}_i+\tilde{y}_i-y_n)^2\\
			=&\sum_{n=n_i+1}^{n_{i+1}}\alpha_n\left[(z-\tilde{y}_i)^2+(\tilde{y}_i-y_n)^2\right].\nonumber
		\end{align}
		Hence,
		\begin{align}
			\argmin_{z\in\mathcal Q}\sum_{n=n_i+1}^{n_{i+1}}\alpha_n(z-y_n)^2=&\argmin_{z\in\mathcal Q}\sum_{n=n_i+1}^{n_{i+1}}\alpha_n(z-\tilde{y_i})^2\nonumber\\
			=&\argmin_{z\in\mathcal Q}(z-\tilde{y}_i)^2
		\end{align}
		which concludes the proof.
		\end{proof}
	\end{proposition}

	\begin{proposition}\label{thm:monotonequant}
		We have
		\begin{align}
			\argmin_{z\in\mathcal{Q}}\left|z-\tilde{y}_i\right|\leq \argmin_{z\in\mathcal{Q}}\left|z-\tilde{y}_j\right|&&\text{if }\tilde{y}_i\leq\tilde{y}_j.	
		\end{align}
	\begin{proof}
		The proof comes from the triangle inequality.
	\end{proof}
	\end{proposition}
	
	\begin{theorem}\label{thm:staircase}
		There exists a minimizer $C_N(\cdot)$ for \autoref{def:isoquant} with $I$ groups such that (with an abuse of notation on $C_N(\cdot)$)
		\begin{align*}
			C_N(\{x_n\}_{n=n_i+1}^{n_{i+1}})=\argmin_{z\in\mathcal Q}|z-\tilde{y}_i|, &&\tilde{y}_i\leq \tilde{y}_{i+1}, \forall i.
		\end{align*}
		where $\tilde{y_i}$ are the weighted group averages.
		\begin{proof}
			The proof comes from \autoref{thm:group}, \autoref{thm:quant} and \autoref{thm:monotonequant}.
		\end{proof}
	\end{theorem}
	
\subsection{Linear Time Batch Algorithm}\label{sec:offline}
	We have shown that an optimal quantized monotone mapping can be derived from an optimal monotone mapping. Hence, we can utilize the following linear time algorithm.
	
\begin{enumerate}
	\item Let us have $N$ samples with the scores $\{x_n\}_{n=1}^N$, where $x_n$ are ordered.
	\item Create a group for each sample, i.e., $n\in\{1,\ldots,N\}$. Set the initial minimizers $\tilde{y}_n=y_n$  and the group weights $\tilde{\alpha}_n=\alpha_n$ for all $n\in\{1,\ldots,N\}$ groups.
	\item Let $\mathcal{I}$ be the set of all indices $i$ such that the consecutive group pair $(i,i+1)$ have $\tilde{y}_i\geq \tilde{y}_{i+1}$\label{step:checkE}
	\item IF $\mathcal{I}$ is not empty, join the groups $i$ and $i+1$ for all $i\in\mathcal{I}$. If there is consecutive grouping, join all of them; e.g., if the pairs $\{i,i+1\}$, $\{i+1,i+2\}$ are joined, we join $\{i,i+1,i+2\}$. 
	When combining the groups $\{i,i+1,\ldots,i+j\}$, we have the new group minimizer $\tilde{y}=\sum_{i'=i}^{j}\tilde{\alpha}_i\tilde{y}_i/\sum_{i'=i}^{j}\tilde{\alpha}_i$ and weight $\tilde{\alpha}=\sum_{i'=i}^{j}\tilde{\alpha}_i$. Re-index the groups accordingly and return to Step \ref{step:checkE}.
	\item ELSE (if $\mathcal{I}$ is empty), project all the group minimizers $\tilde{y}_i$ on to the set $\mathcal{Q}$ to create an optimal quantized mapping.
\end{enumerate}

\begin{remark}
	It takes $O(1)$ time to update the weighted means and the weight sums per group involved in the joining. Thus, this approach can find an optimal transform in linear space and linear time.
\end{remark}

\section{Sequentializing Isotonic L2 Regression}\label{sec:prefix}

We first consider the easier version of the sequential problem. Let the samples come in an ordered fashion such that $x_{n+1}\geq x_{n}$, where the goal is to sequentially update an optimal mapping $C_N(\cdot)$ with the new sample $x_{N+1}$. An optimal mapping for the first $N$ samples are given by
\begin{align*}
	\sum_{n=1}^N\alpha_n(C_N(x_n)-y_n)^2=\min_{C(\cdot)\in\Omega_Q}\sum_{n=1}^N\alpha_n(C(x_n)-y_n)^2,
\end{align*}
To create $C_{N+1}(\cdot)$ using $x_{N+1}$, we can utilize the standard batch algorithm in \autoref{sec:offline}, since the joining order of the groups does not matter. The sequential algorithm is as follows:

\begin{enumerate}
	\item We have $N=1$, $x_1$, the minimizer $\tilde{y}_1=y_1$, the weight $\tilde{\alpha}_1=\alpha_1$ and the mapping $\tilde{z}_1=\argmin_{z\in\mathcal Q}|z-\tilde{y}_1|$.
	\item Let the number of groups be $I-1$. Receive $x_{N+1}$; set $\tilde{y}_I=y_{N+1}$, $\tilde{\alpha}_I=\alpha_{N+1}$, $\tilde{z}_I=\argmin_{z\in\mathcal Q}|z-\tilde{y}_I|$. \label{step:checkO}
	\item WHILE the group with the largest index (denoted as $I$) needs to be joined with its immediate preceding group, join them. Set $\tilde{y}=(\tilde{\alpha}_{I-1}\tilde{y}_{I-1}+\tilde{\alpha}_{I}\tilde{y}_I)/(\tilde{\alpha}_{I-1}+\tilde{\alpha}_{I})$, $\tilde{z}=\argmin_{z\in\mathcal Q}|z-\tilde{y}|$ and $\tilde{\alpha}=\tilde{\alpha}_{I-1}+\tilde{\alpha}_I$. Re-index the new group with the preceding group's index $I-1$.\label{step:join}
	\item Set $N\leftarrow N+1$ and return to Step \ref{step:checkO}.
\end{enumerate}

For a given time $T$, let $I_T$ be the number of groups in the mapping. Since with each joining the number of groups decreases by $1$, we will have exactly $T-I_T$ joining in total. Since each joining takes $O(1)$ time, the worst-case time complexity is $O(T)$. Note that while the cumulative time complexity of this approach is $O(T)$, the per sample time complexity by itself is also $O(T)$ since the decrease in the number of steps is only bounded by the number of groups. Hence, we can find an optimal mapping in linear space and linear amortized time. 
To limit the per sample time complexity, we can utilize the quantized structure of the mapping.

\begin{lemma}\label{thm:samequant}
	If two consecutive groups have the same quantized mapping, they will have so in the future.
	\begin{proof}
		Let the groups $i$ and $i+1$ have the group minimizers $\tilde{y}_i$, $\tilde{y}_{i+1}$ together with their quantizations $\tilde{z}_i$ and $\tilde{z}_{i+1}$. For the quantized mappings to change, their minimizers should change. For $\tilde{y}_i$ to change, $\tilde{y}_{i+1}$ should decrease. If the change in $\tilde{y}_{i+1}$ is not significant enough, $\tilde{z}_{i+1}$ will not change. However, if it is so, $\tilde{y}_{i+1}$ will decrease beyond $\tilde{y}_i$ and the $(i+1)^{th}$ group will need to be combined with the $i^{th}$ group. Hence, their minimizers and their respective quantized mapping will be the same. Thus, if the quantized mappings of two consecutive groups are the same, it will always be so.
	\end{proof}
\end{lemma}

\begin{theorem}\label{thm:joinquant}
	We can combine a consecutive group pair $(i,i+1)$ not only when $\tilde{y}_i\geq \tilde{y}_{i+1}$ but also $\tilde{z}_i=\tilde{z}_{i+1}$.
	\begin{proof}
		The proof comes from \autoref{thm:samequant}.
	\end{proof}
\end{theorem}

By utilizing \autoref{thm:joinquant}, we can also consider the consecutive groups with the same quantized mappings to be joined. This will substantially decrease the per time complexity, since the number of groups will be limited by the quantization. If the mapped set $\mathcal{Q}$ is bounded it will be constant per time, otherwise the complexity will be pseudo-linear since it will depend on the observed samples so far.

\section{A Log-Linear Time Truly Sequential Algorithm}\label{sec:recursive}
To create the mappings $C_N(\cdot)$ when the sample scores arrive out of order, we propose a recursive algorithm that iteratively merges sets. Our algorithm has a sequential implementation with logarithmic time complexity per sample. We first provide a batch optimization version to increase comprehension.

At the first stage (i.e., the bottom level), for $n\in\{1,\ldots,N\}$, we have the mapping set $\A^{1,n}=\{\Y^{1,n},\Z^{1,n}\}$ where $\Y^{1,n}=\{y_n\}$ are the target variables, $\Z^{1,n}=\{z_n\}$ are the optimal quantized mappings and the indices $n$ are implied by the ordinality of the scores $\X=\{x_n\}_{n=1}^N$.
Starting from the bottom level $k=1$ (initial stage), at every level $k$, we create the sets $\A^{k+1,j}$ by merging the adjacent relevant sets at the $k^{th}$ level. Whenever the sets $\A^{k,i}=\{\Y^{k,i},\Z^{k,i}\}$ and $\A^{k,i+1}=\{\Y^{k,i+1},\Z^{k,i+1}\}$ are merged together, we have the new target variable set $\Y^{k+1,j}=\Y^{k,i}\cup\Y^{k,i+1}$, and the new quantized mappings $\Z^{k+1,j}$, which can be calculated in time linear with the cardinality $|\Y^{k,i+1}|$. If there is no other set that can be merged with $\A^{k,i}$, it is moved up a level, i.e., $\A^{k+1,j}=\A^{k,i}$. Note that $i$ may not be twice of $j$; $i$, $j$ are the relative indices at their respective levels. Since the combinations are done via a binary tree, the total time complexity will be log-linear, i.e., $O(N\log N)$ and the space complexity is linear $O(N)$. 

When the samples sequentially arrive in an unordered manner, we can run this merging algorithm from the stretch with each new sample to create an optimal quantized mapping. However, such an implementation would be inefficient and take linear time complexity per sample. To this end, we modify this binary merging structure to create an efficient sequential algorithm. First of all to efficiently combine the mappings of different sets, we utilize auxiliary variables. We modify the set $\mathcal{A}^{k,i}$ associated with the index $i$ at the level $k$ as the following
\begin{align}
	\mathcal{A}^{k,i}=\{\mathcal{X}_0^{k,i},\mathcal{X}_1^{k,i},\mathcal{Y}^{k,i},\mathcal{W}^{k,i}\},
\end{align}
where $|\mathcal{X}_0^{k,i}|=|\mathcal{X}_1^{k,i}|=|\mathcal{Y}^{k,i}|=|\mathcal{W}^{k,i}|=I_{k,i}$ is the number of groups in the mapping of $(k,i)$; $\mathcal{X}_0^{k,i}$ are the smallest scores in each group, $\mathcal{X}_1^{k,i}$ are the largest scores in each group, $\mathcal{Y}^{k,i}$ are the group minimizers and $\mathcal{W}^{k,i}$ are the total group weights.

\begin{enumerate}
	\item At any point in the algorithm, whenever a new sample comes, we only update the necessary intermediate sets. 
	\item Schedule a new set $\A_m$ to move up as itself to the next level if it does not fall between a pair of already-combined sets at a level $k$. .
	\item If a new set $\A_m$ (middle) falls between a pair of already-combined sets $\A_l$ (left) and $\A_r$ (right) at a level $k$, combine $\A_l$ with $\A_m$ instead of $\A_r$ (left-biased), i.e., the preceding merger of $\A_l$ with $\A_r$ is updated with the new merger of $\A_l$ with $\A_m$. Schedule $\A_r$ to move up as itself to the next level.
	\item If a scheduled-to-move-up set is adjacent to an already-moved-up set at any level $k$, update the moved-up set with their merger at the next level. If not, move up scheduled-to-move-up set as itself to the next level $k+1$.
	\item Whenever a set in the recursive structure is updated (a new or updated merger) at a level $k$, update the subsequent mergers at the next level $k+1$.
\end{enumerate}

\subsection{Complexity analysis}\label{sec:complexity}

\begin{lemma}\label{lem:depth}
	The depth of the recursion tree is logarithmic in the number of samples $N$, i.e., $D=O(\log N)$ (where $k\leq D$, $\forall k$), and has size $O(N)$.
	\begin{proof}
		The structure of the sequential algorithm forces any two adjacent not-combined sets at any level to be combined at the next level. Let the number of sets that exists at the level $k$ be $N_k$, then the number of sets at the level $k+1$ is bounded as $N_{k+1}+1\leq \frac{2}{3}(N_{k}+2)$ and concludes the proof.
	\end{proof}
\end{lemma}

\begin{lemma}\label{lem:traverse}
	The number of sets that are updated with each sample is logarithmic in the number of samples $N$, i.e., $O(\log N)$.
	\begin{proof}
		An updated set at the level $k+1$ will result from a preceding connected set update at the level $k$. For any new sample, we will have at most two updates at each level and the updates in different levels are connected with chains. Since the depth of the recursion is logarithmic, the number of updates are logarithmic in $N$, which concludes the proof.
	\end{proof} 
\end{lemma}

\begin{lemma}\label{lem:update}
	Let us have two adjacent sets $\A^{0}$ and $\A^{1}$, where $\mathcal{A}^0=\{\mathcal{X}_0^0,\mathcal{X}_1^0,\mathcal{Y}^0,\mathcal{W}^0\}$ and $\mathcal{A}^1=\{\mathcal{X}_0^1,\mathcal{X}_1^1,\mathcal{Y}^1,\mathcal{W}^1\}$ with the corresponding number of groups $I_0$ and $I_1$. The combination of $\mathcal{A}^0$ and $\mathcal{A}^1$ denoted by $\mathcal{A}=\{\mathcal{X}_0,\mathcal{X}_1,\mathcal{Y},\mathcal{W}\}$ with the corresponding number of groups $I$ can be calculated in $O(I_1+I_0-I)$ time complexity.
	\begin{proof}
		Because of the monotone mapping structure $I\leq I_0+I_1$. By utilizing the iterative pairwise group combination in the algorithm of \autoref{sec:prefix} with the group minimizers and the group weights, we can decrease the number of groups in the combination by $1$ in $O(1)$ time. Note that we not only combine the groups when adjacent group minimizers violate monotonicity but also when the quantized mappings are the same as well. Hence, the total time complexity will be $O(I_1+I_0-I)$.
	\end{proof}
\end{lemma}

\begin{theorem}\label{thm:complexity}
	The sequential algorithm updates the optimal mapping in $O(K\log N)$ time per sample for some $K$, which is the upper bound of the number of groups in the sets.
	\begin{proof}
		With a new sample, the number of updates totals to $O(\log N)$ from \autoref*{lem:depth} and \autoref*{lem:traverse}. Since each update takes a time that is linear with the number of groups from \autoref*{lem:update}, the mapping is updated in $O(K\log N)$ time, where $K$ is an upper bound on the number of groups.
	\end{proof}
\end{theorem}

\begin{remark}
	In the traditional isotonic L2 regression (which is done on the set of reals), the number of groups will be bounded by $N$; hence, the per sample time complexity would be linear. On the other hand, when the quantization set $\mathcal{Q}$ is bounded with fixed cardinality (e.g., percentage), the per sample time complexity will be logarithmic. If the quantization set is not bounded, then the per sample complexity will be pseudo-linear (dependent on the range of target variables $y_n$).
\end{remark}

Henceforth, we have shown that for isotonic L2 regression problems, it is possible to achieve efficient optimal sequential algorithms, especially when the mapping is quantized.

\bibliographystyle{ieeetran}
\bibliography{double_bib}
\end{document}